\documentclass[conference]{IEEEtran}
\usepackage{times}

\usepackage[numbers]{natbib}
\usepackage{multicol}
\usepackage[bookmarks=true]{hyperref}

\usepackage{amsfonts}
\usepackage{amsmath,bm}
\usepackage{amssymb}
\usepackage{CJK}
\usepackage{indentfirst}
\usepackage{cases}
\usepackage{algorithm}
\usepackage[algo2e]{algorithm2e}
\usepackage{graphicx}
\usepackage{stfloats}
\usepackage{subfigure}
\usepackage{booktabs}
\usepackage{color,soul} 
\definecolor{red}{rgb}{1.00,0.00,0.00}
\definecolor{blue}{rgb}{0.00,0.00,1.00}
\definecolor{green}{rgb}{0.30, 0.50,0.00}

\usepackage{flushend} 
\usepackage{arydshln}
\usepackage{color}
\usepackage{tikz}

\begin{document}

\title{\huge IPPO: Obstacle Avoidance for Robotic Manipulators in Joint Space via Improved Proximal Policy Optimization}



\author{\authorblockN{Yongliang Wang}
\authorblockA{Department of Artificial Intelligence\\
Bernoulli Institute, Faculty of Science and Engineering\\
University of Groningen, The Netherlands\\
Email: yongliang.wang@rug.nl}
\and
\authorblockN{Hamidreza Kasaei}
\authorblockA{Department of Artificial Intelligence\\
Bernoulli Institute, Faculty of Science and Engineering\\
University of Groningen, The Netherlands\\
Email: hamidreza.kasaei@rug.nl}
}


%

\maketitle

\begin{abstract}
Reaching tasks with random targets and obstacles is a challenging task for robotic manipulators. In this study, we propose a novel model-free reinforcement learning approach based on proximal policy optimization (PPO) for training a deep policy to map the task space to the joint space of a 6-DoF manipulator. To facilitate the training process in a large workspace, we develop an efficient representation of environmental inputs and outputs. The calculation of the distance between obstacles and manipulator links is incorporated into the state representation using a geometry-based method. Additionally, to enhance the performance of the model in reaching tasks, we introduce the action ensembles method and design the policy to directly participate in value function updates in PPO. To overcome the challenges associated with training in real-robot environments, we develop a simulation environment in Gazebo to train the model as it produces a smaller Sim-to-Real gap compared to other simulators. However, training in Gazebo is time-intensive. To address this issue, we propose a Sim-to-Sim method to significantly reduce the training time. The trained model is then directly applied in a real-robot setup without fine-tuning. To evaluate the performance of the proposed approach, we perform several rounds of experiments in both simulated and real robots. We also compare the performance of the proposed approach with six baselines. The experimental results demonstrate the effectiveness of the proposed method in performing reaching tasks with and without obstacles. our method outperformed the selected baselines by a large margin in different reaching task scenarios. A video of these experiments has been attached to the paper as supplementary material.

\end{abstract}

\IEEEpeerreviewmaketitle

\section{Introduction}

\begin{figure}[!t]
      \centering
      \includegraphics[width=1.0\linewidth]{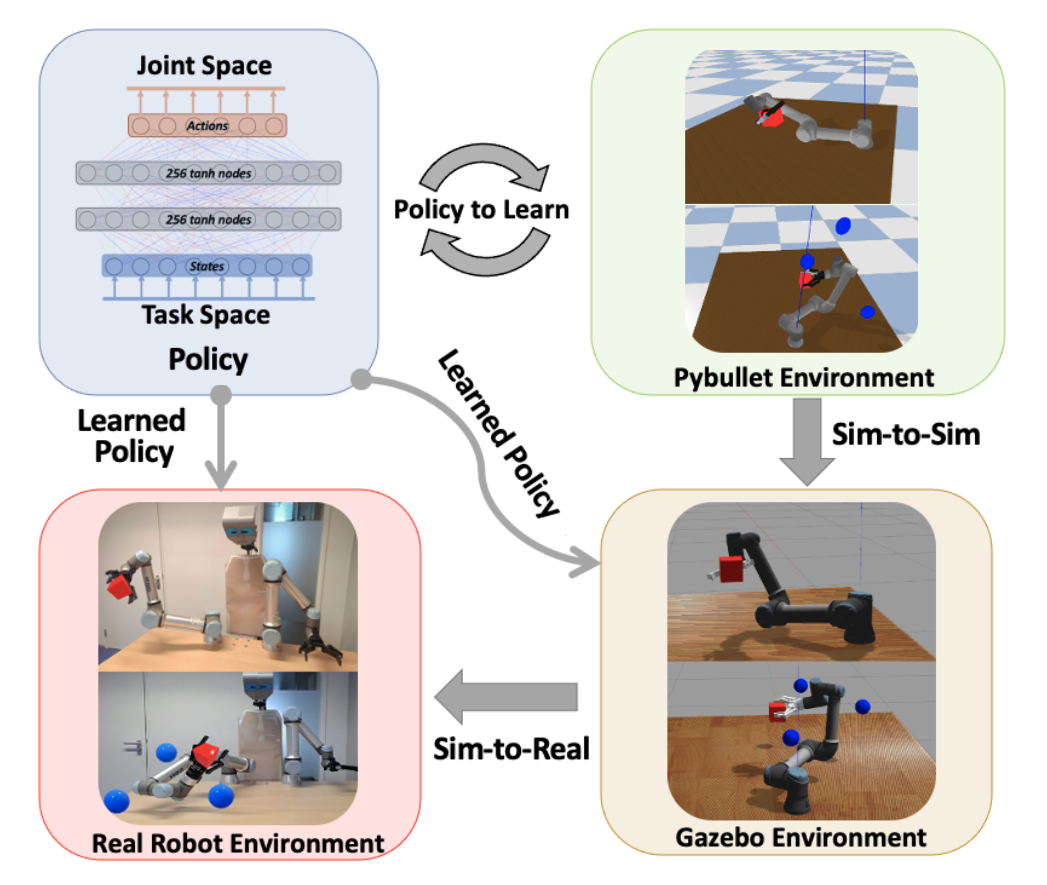}
      \vspace{-4mm}
      \caption{Overview of the proposed approach (IPPO): the agent is first trained in the Pybullet environment to learn the mapping from task space to the joint space for the UR5e manipulator. The trained policy is then used in Gazebo for the Sim-to-Sim adaptation, followed by direct application in real robot scenarios without further fine-tuning. }
      \label{figure1}
\end{figure}
The goal-reaching task is a crucial capability for robotic manipulators, as it is a fundamental requirement for many robotic applications~\cite{chen2022deep, sun2022fully, rudin2022learning, liu2022digital, matsuzaki2022learning}. In human-centric environments, robots often operate in complex and dynamic domains where the presence of obstacles makes motion planning a challenging task. The motion-planning task for high-degree-of-freedom robotic manipulators in dynamic environments is challenging due to the need for a mathematical model that is both complex and difficult to establish. Classical approaches, such as the rapid exploring random trees (RRT) algorithm~\cite{karaman2011anytime} and the node control-bidirectional RRT (NC-BRRT) algorithm~\cite{10011642}, have limitations in handling dynamic environments and often require prior knowledge of the surroundings, leading to intensive computation. Previous research has shown that such motion planning methods are inadequate for dynamic domains~\cite{chen2022deep, kunz2010real, xu2021motion}, leading to a need for the development of advanced methods that can effectively handle these challenges.




Linghuan $et$ $al.$ \cite{8681733} proposed an adaptive neural network bounded control scheme for an n-link rigid robotic manipulator with unknown dynamics. However, their method necessitated prior knowledge of environmental limitations. Current path-planning approaches for dynamic environments relied on having prior knowledge of the surroundings and required intense online computation \cite{9151285, mnih2015human}. When the environment was complex, the excessive online computation could make the system unresponsive to change in its state. To sum up, when the targets and obstacles change at random, the motion planning task for high-degree-of-freedom manipulators will become notoriously challenging in such uncertain environments as mathematical models that are complex and difficult to establish~\cite{aljalbout2021learning}. Meanwhile, traditional control approaches are often unable to navigate in unstructured environments. Besides, most methods rely on solving inverse kinematics or dynamics equations to map the task space to joint space for collision avoidance, which requires a highly accurate model of the robot's dynamics and is not easily generalized to diverse tasks and different robots~\cite{zhou2021robotic, rodriguez2006obstacle}. 

 In recent years, deep Reinforcement Learning (RL) offers a promising solution to this challenge by providing a data-driven approach to motion planning~\cite{yang2022abstract}. RL algorithms can learn from interactions with the environment, enabling robots to perform complex goal-reaching tasks while avoiding obstacles in real-time. RL-based methods for robotic manipulation have gained popularity and are increasingly being used as an alternative to traditional analytical control systems~\cite{kober2013reinforcement, zhu2017target, jauhri2022robot, niko, CRUZ201734, 9205217}, as they have shown great potential for improving the accuracy and efficiency of goal-reaching tasks. For instance, Adel $et$ $al.$ \cite{baselizadeh2022motion} proposed a reinforcement learning framework that combines nonlinear model predictive control with obstacle avoidance. This approach was evaluated on a 6-DoF robot manipulator and demonstrated its ability to successfully avoid collisions with static obstacles. However, it has limitations when it comes to handling obstacles in dynamic environments. Research has shown that the presence of moving obstacles can greatly increase the difficulty of motion planning tasks~\cite{quan2023obstacle}. 

Adarsh Sehgal $et$ $al.$ proposed a deep deterministic policy gradient (DDPG) and hindsight experience replay (HER) based method using of the genetic algorithm (GA) to fine-tune the parameters’ values. They experimented on six robotic manipulation tasks and got better results than baselines \cite{sehgal2022automatic}. Franceschetti $et$ $al.$ proposed an extensive comparison of the trust region policy optimization (TRPO) and deep Q-Network with normalized advantage functions (DQN-NAF) with respect to other state-of-the-art algorithms, namely DDPG and vanilla policy gradient (VPG)~\cite{franceschetti2022robotic}. Unlike our work, these studies only concentrate on reaching a single target position. For multi-target trajectory planning, Wang $et$ $al.$ introduced action ensembles based on the Poisson distribution (AEP) to PPO, their method could be easily extended to realize the task that the end-effector tracks a specific trajectory \cite{9636681}. For space robots, the workspace is enough to complete tasks, but for industrial robots, it is insufficient. Their approach did not produce favorable outcomes when applied to a larger workspace after training. Thus, the algorithm requires further development. In another work, Kumar $et$ $al.$ proposed a simple, versatile joint-level controller via PPO. Experiments showed the method capable of achieving similar error to traditional methods, while greatly simplifying the process by automatically handling redundancy, joint limits, and acceleration or deceleration profiles \cite{kumar2021joint}. Nevertheless, the output of the neural network is the velocity of the end-effector. Additionally, the majority of DRL-based research completes learning in task space rather than joint space \cite{wang2022end, zuo2022graph, borja2022affordance}, which is prone to produce weak results for reaching tasks \cite{hansen2022visuotactile, wang2022end}. Furthermore, such approaches still need to calculate the inverse kinematics and cannot accomplish reaching tasks when obstacles are close to the manipulator's links. 

In this paper, we propose a novel model-free deep reinforcement learning approach, called Improved PPO (IPPO), to tackle reaching multi-target goals while avoiding obstacles in dynamic environments. An overview of the method is depicted in Fig.~\ref{figure1}. In particular, we train a deep policy to map from task space to joint space for a 6-DoF manipulator. To improve the effectiveness of the model's output on reaching tasks, the action ensembles method is introduced, and the policy is designed to join in value function updates directly in PPO. Additionally, since training such a task in real-robot is time-consuming and strenuous, we develop a simulation environment to train the model in Gazebo as it produces a smaller Sim-to-Real gap compared to other simulators. However, as training robots in Gazebo is computationally expensive and requires a long training time, we propose a Sim-to-Sim method to significantly reduce the training time. Finally, the trained model is directly used in a real-robot setup without fine-tuning. 

 In comparison to prior works \cite{gu2022anti, hsu2022improving, sadhukhan2021multi, kobayashi2021proximal}, our research makes three significant contributions: (\textit{i}) The calculation of the distance between obstacles and the manipulator's links is done using a geometry-based method, which improves the reaching task in the presence of obstacles. (\textit{ii}) An action ensembles approach is introduced to enhance the efficiency of the policy. (\textit{iii}) An adaptive discount factor for PPO is designed, allowing the policy to directly participate in the value function update. Empirical results demonstrate that the proposed approach outperforms other baseline methods in different testing scenarios.

\section{Preliminary}

\begin{figure}[!b]
      \centering
      \includegraphics[width=1.0\linewidth]{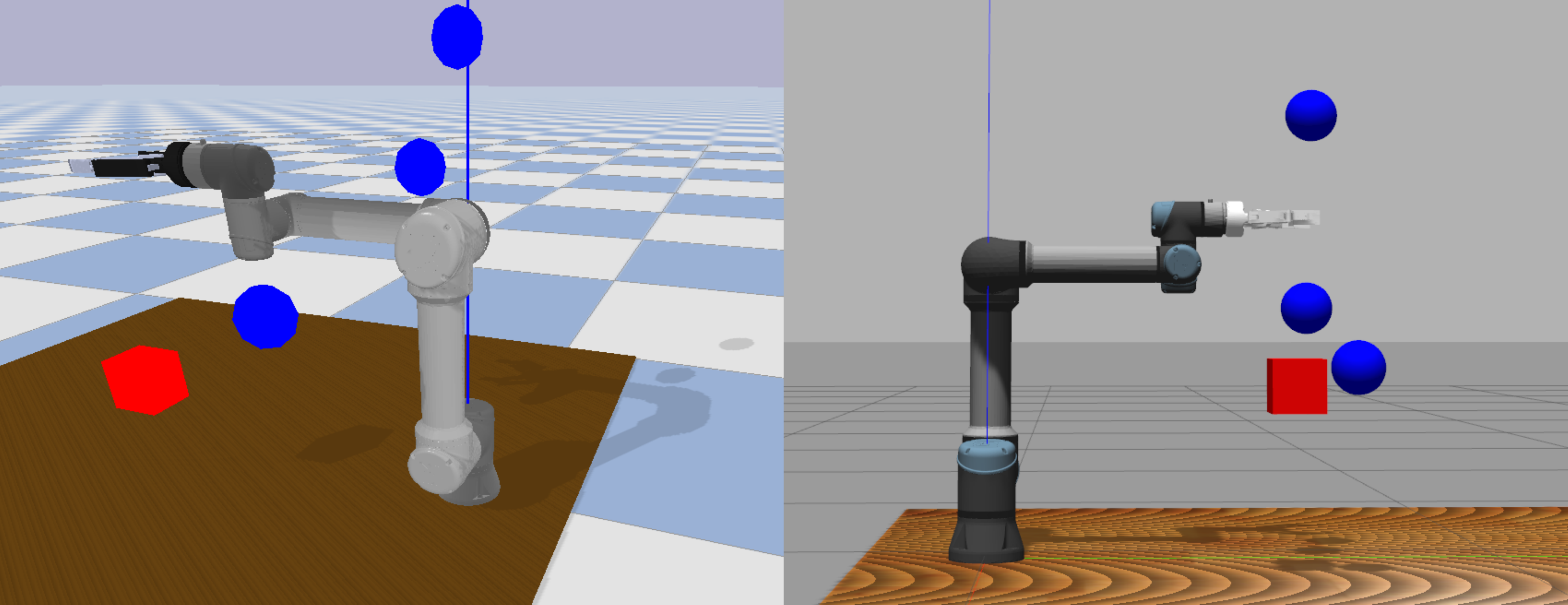}
      \vspace{-3mm}
      \caption{The environment of 6-DoF manipulator in the Pybullet (\textit{left}) and the Gazebo (\textit{right}): The goal is shown by a red block and obstacles are highlighted by blue spheres. }
      \label{figure2}
  \end{figure}

Our research focuses on developing an effective method for obstacle avoidance in reaching tasks for manipulators. To achieve this goal, we aim to make the manipulator safely interact with the environment multiple times. To design our training process, we initially chose Gazebo due to its better compatibility with the Robot Operating System (ROS) compared to Pybullet. However, DRL methods tend to have a long training time in Gazebo. To mitigate this challenge, we created a similar environment in Pybullet for initial training, and then transferred and evaluated the learned model in Gazebo through a Sim-to-Sim transfer process. Finally, we tested the efficiency of the model on a real robot using Sim-to-Real transfer. The simulation environments in Gazebo and Pybullet are depicted in Fig.~\ref{figure2}. In both environments, a UR5e robot equipped with a robotiq\_$140$ is utilized as the manipulator. During the training and testing phases, the pose of the target (represented in red) and obstacles (represented in blue) is randomly set within the workspace.

\subsection{Proximal Policy Optimization (PPO)}

In this work, we use PPO, one of the state-of-the-art online DRL methods, since it is known for its stability and effectiveness in various environments. The algorithm balances exploration and exploitation to find a policy that maximizes the reward. PPO also offers a trade-off between stability and high sample efficiency, making it suitable for the complex robotic environment with obstacle avoidance. 
In particular, PPO is a type of policy gradient training that alternates between sampling data through environmental interaction and optimizing a clipped surrogate objective function using stochastic gradient descent \cite{schulman2017proximal}. The clipped surrogate objective function improves training stability by limiting the size of the policy change at each step. In PPO, the clipped surrogate objective function is designed as follows:
\begin{equation}
   \begin{aligned}
    \bm{L}^{\bm{CLIP}}(\theta^{\pi})&=\mathbb{\dot{E}}_t[min(r_t(\theta^{\pi})\hat{A_t},\\
    &\quad clip(r_t(\theta^{\pi}),1-\epsilon,1+\epsilon)\hat{A_t})]
   \end{aligned}
\end{equation}
\begin{equation}
    \hat{A_t} = \delta_t+(\gamma\lambda)\delta_{t+1}+\cdots+\cdots+(\gamma\lambda)^{T-t+1}\delta_{T-1}
\end{equation}
\begin{equation}
    \delta_t = r_t+\gamma V(s_{t+1})-V(s_t)
\end{equation}
\begin{equation}
    V(s) = \mathbb{E}_{s,a\sim\pi}[G(s)|s]
\end{equation}
\begin{equation}
    G(s) = \sum_{i=t}^{\infty}\gamma ^ {i-t}r(s_i)
\end{equation}
where $\theta^{\pi}$ is the parameters of the policy neural network, $r_t(\theta^{\pi})$ denotes the probability ratio, $\hat{A_t}$ represents the generalized advantage estimator (GAE) and is used to calculate the policy gradient. The reward value at $t$ is shown by $r_t$, and the $\epsilon$ is a constant between 0 and 1, which is set to 0.2 in the baseline algorithm. $\gamma = 0.99$,  $V(s)$ refers to the expected return of state $s$ and $G$ represents the discounted cumulative reward. Likewise, $V_{target}(s)$ is the target value. Additionally, the value loss function is expressed as follows:
\begin{equation}
    \bm{L}_V(\theta^{V})=\mathbb{E}_{s,a\sim\pi}[(V(s)-V_{target}(s))^ 2]
\end{equation}

\subsection{Sim-to-Sim and Sim-to-Real Adaptation}

As stated earlier, we train a DRL policy in Pybullet first and then transfer it to Gazebo to reduce training time~\cite{pinosky2022hybrid, ju2022transferring, sharma2022learning}. The final goal is to evaluate the policy's performance in a real-world scenario through a Sim-to-Real adaptation~\cite{scheikl2022sim, martin2022pre}. While previous research in the field of learning navigation and manipulation policies have focused on bridging the Sim-to-Real gap in domain adaptation~\cite{zhu2017target, zhang2021sim2real, bousmalis2018using, fang2018multi}, our work differs in that obstacle avoidance is accomplished in joint space. Our primary objective is to achieve high accuracy in simulation for real-world applications. The Sim-to-Sim transfer method is efficient in reducing training time and evaluating the robustness of the proposed approach over noises and inaccurate robot models before deploying the learned model on a real-robot platform. Therefore,  
we consider both Sim-to-Sim and Sim-to-Real transfers~\cite{scheikl2022sim, chen2022zero, kadokawa2022cyclic} in order to quickly train and evaluate the proposed model in a variety of tasks and domains. As illustrated in Fig.~\ref{figure1}, we train a deep policy in the Pybullet environment to learn the mapping from the task space to the joint space of the UR5e manipulator. The learned policy is then subjected to a Sim-to-Sim phase in the Gazebo environment, allowing us to evaluate and test the policy within a simulated environment prior to deployment in real-world scenarios. Finally, the policy is directly applied in real-robot without further refinement or fine-tuning. 

\section{Strategy for Learning}
We adopt PPO to accomplish obstacle avoidance with the mapping from task space to joint space. For reinforcement learning, one of the important aspects is to devise a good learning strategy~\cite{wang2022end, ju2022transferring, li2022towards}, which includes selecting an appropriate state and action representation \cite{thumm2022provably, wong2022oscar, elguea2023review}. The strategy is implemented as a deep policy, which is designed as a multi-layer perception network with two hidden layers.

\subsection{State and Action Representations}

It is crucial for DRL methods to choose the appropriate state and action space. Most researchers prefer to represent both states and actions in task space, which is ineffective for avoiding collision between links and obstacles. To accomplish collision avoidance in the whole workspace, we consider the position of 6 joints, end-effector, and targets as part of state representation. Furthermore, the  errors in X, Y, and Z axes, and the distance between obstacles and the five links are also considered state representation. It is worth mentioning that we do not consider the distance of the obstacles to the base link. Therefore, the state is represented as a vector: $s \in \mathbb{R}^{19}$. For action representation, we consider the position of the six joints in order to avoid complex and time-consuming inverse kinematics calculations and map from task space to joint space. In the following subsections, we discuss the state and action spaces in more detail. 

\subsubsection{States in Reaching Task without Obstacles}
In the case of the obstacle-free goal-reaching task, we represent the state as:
\begin{equation}
    \bm s_t = <\bm q_t, \bm p_e, \bm p_t, \bm {error}>
\end{equation} 
where $\bm q_t = (q_{t1} \dots q_{t6})$ is the position of the six joints, $\bm p_e = (p_{ex}, p_{ey}, p_{ez})$ represents the position of the end-effector, $\bm p_t = (p_{tx}, p_{ty}, p_{tz})$ is referred to the target position. $\bm {error} = (e, e_x, e_y, e_z)$ is the error vector including absolute distance and distances in X, Y, and Z axes, respectively.   

\subsubsection{States in Reaching Task with Obstacles} 
\begin{figure}[!t]
      \vspace{-3mm}
      \centering
       \begin{minipage}[t]{1.0\linewidth}
        \centering
        \includegraphics[width=0.45\linewidth]{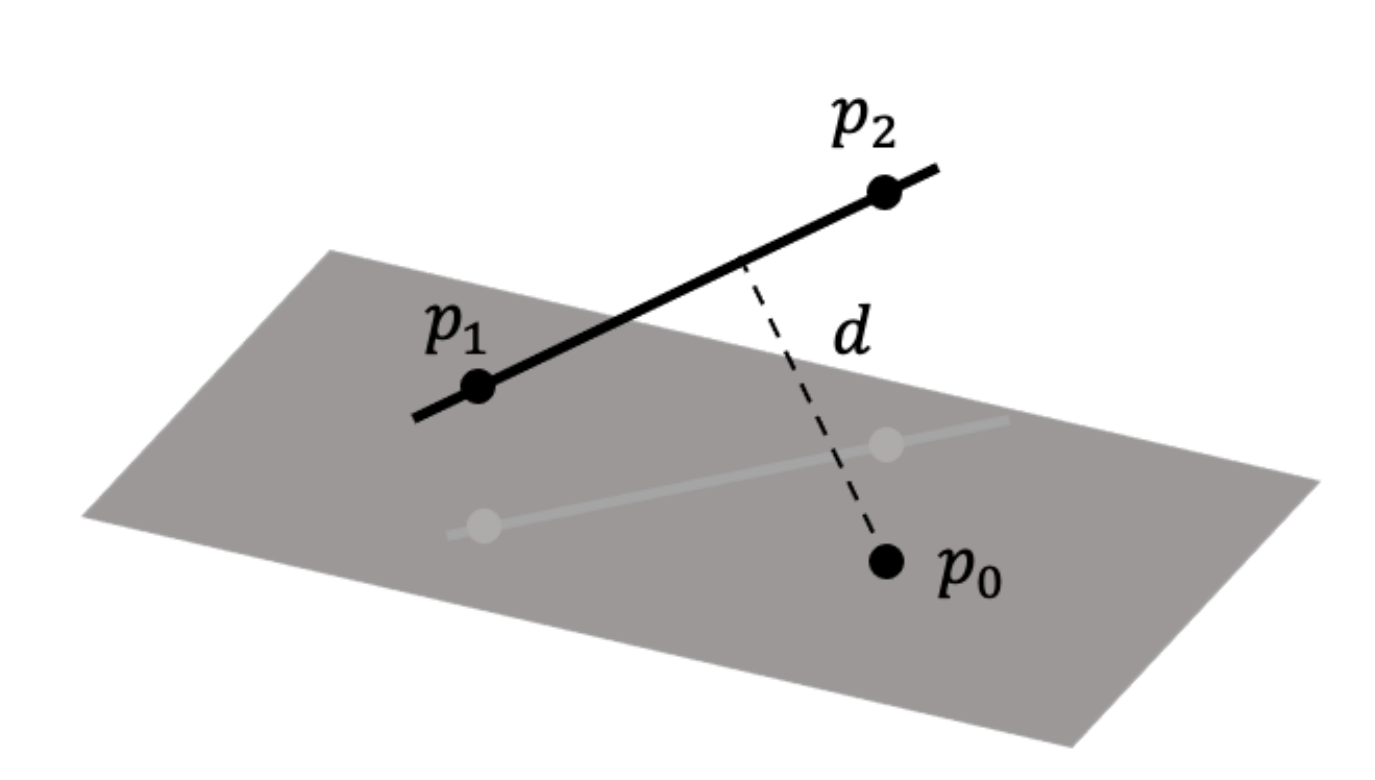}
        \includegraphics[width=0.45\linewidth]{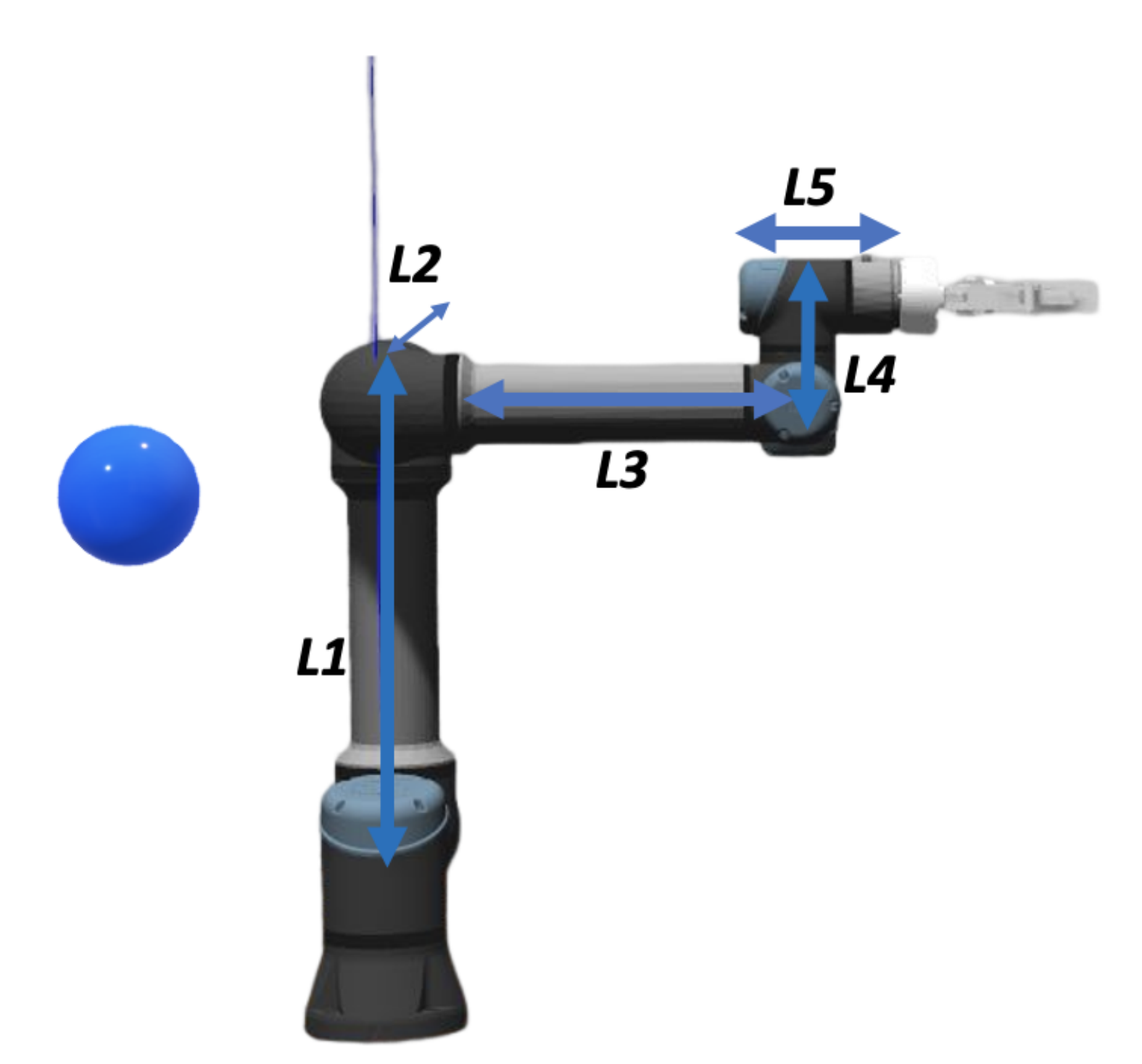}
    \end{minipage}
      \vspace{-4mm}
      \caption{The distance calculation between a point and a line in space (\textit{left}). $\bm d_{obs}$ is obtained by calculating the distance between the obstacle (blue) and 5 links.}
      \label{figure3}
  \end{figure}

When there are obstacles in the environment, the state is represented as:

\begin{equation}
    \bm s_t = <\bm q_t, \bm p_e, \bm p_t, \bm {error}, \bm d_{obs}>
\end{equation} 
where  $\bm d_{obs}$ is the shortest distance between obstacles and links in space. As depicted in Fig.~\ref{figure3}, to calculate the distance between the obstacle and each link in joint space, we transform it into a geometric problem to find the shortest distance between any point in space and different links. We provide an example on the left side of Fig.~\ref{figure3} to make it clearer. Let a three-dimensional line be specified by two points, $p_1 = (x_1, y_1, z_1)$ and $p_2 = (x_2, y_2, z_2)$, where $\cdot$ represents the dot product. Therefore, a vector along the line is given by the following equation:

\begin{equation}
    \bm v = \left [\begin{array}{cccc} x_1+(x_2-x_1)t\\\nonumber
                                       y_1+(y_2-y_1)t\\\nonumber
                                        z_1+(z_2-z_1)t\end{array}\right]
\end{equation} 
The squared distance between a point on the line with parameter $t$ and a point $p_0 = (x_0, y_0, z_0)$ is therefore:

\begin{equation}
   \begin{aligned}
    d^2 = [(x_1-x_0)+(x_2-x_1)t]^2\\
    +[(y_1-y_0)+(y_2-y_1)t]^2\\
    +[(z_1-z_0)+(z_2-z_1)t]^2
   \end{aligned}
\end{equation}
Set $d(d^2)/dt=0$ and solve for $t$ to obtain the shortest distance:
\begin{equation}
    \bm t = -\dfrac{(x_1-x_0)\cdot(x_2-x_1)}{|x_2-x_1|^2}
\end{equation} 
The shortest distance can then be calculated by plugging Eq. (10) back into Eq. (9). Thus, as shown on the right side of Fig.~\ref{figure3}, we consider each link of the robot as the line and the obstacle as the point. We then calculate the shortest distance between every obstacle and link. 

\subsubsection{Actions}
In both cases, the action space is represented by a vector, $\bm a_t = <\dot{\bm q}_t>$, where $\dot{\bm q}_t = (\dot{q}_{t1} \dots \dot{q}_{t6})$ represents the position of the six joints.

\subsection{Reward Function}

The following function represents how we calculate the reward for various situations:
\begin{equation}
    \bm{R}(s,a)= -[\omega_1e^2+\ln{(e^2+\tau_e)}+\omega_2\sum_{i=1}^{n}\psi_i]
\end{equation}
\begin{equation}
    \psi_i=max(0, 1-||d_i||/d_{max})
\end{equation}
where $e=\|p_t-p_e\|$ refers to the euclidean distance between the target pose and the end-effector. The middle term ($\ln(\cdot)$) encourages the end-effector error to tend to be zero, and the $\psi_i$ represents the penalties of obstacle avoidance, 
$\omega_1,$ and $\omega_2$ are two coefficients, and $\tau_e$ represents the threshold of error between the end-effector and the target pose. Based on trial and errors, we set $\omega_1 = 10^{-3}$, $\omega_2 = 0.1$, $d_{max} = 0.05$ and $\tau_e = 10^{-4}$. 

\subsection{Neural Network Structure}

\begin{figure}[!b]
      \centering
      \includegraphics[width=1.0\linewidth]{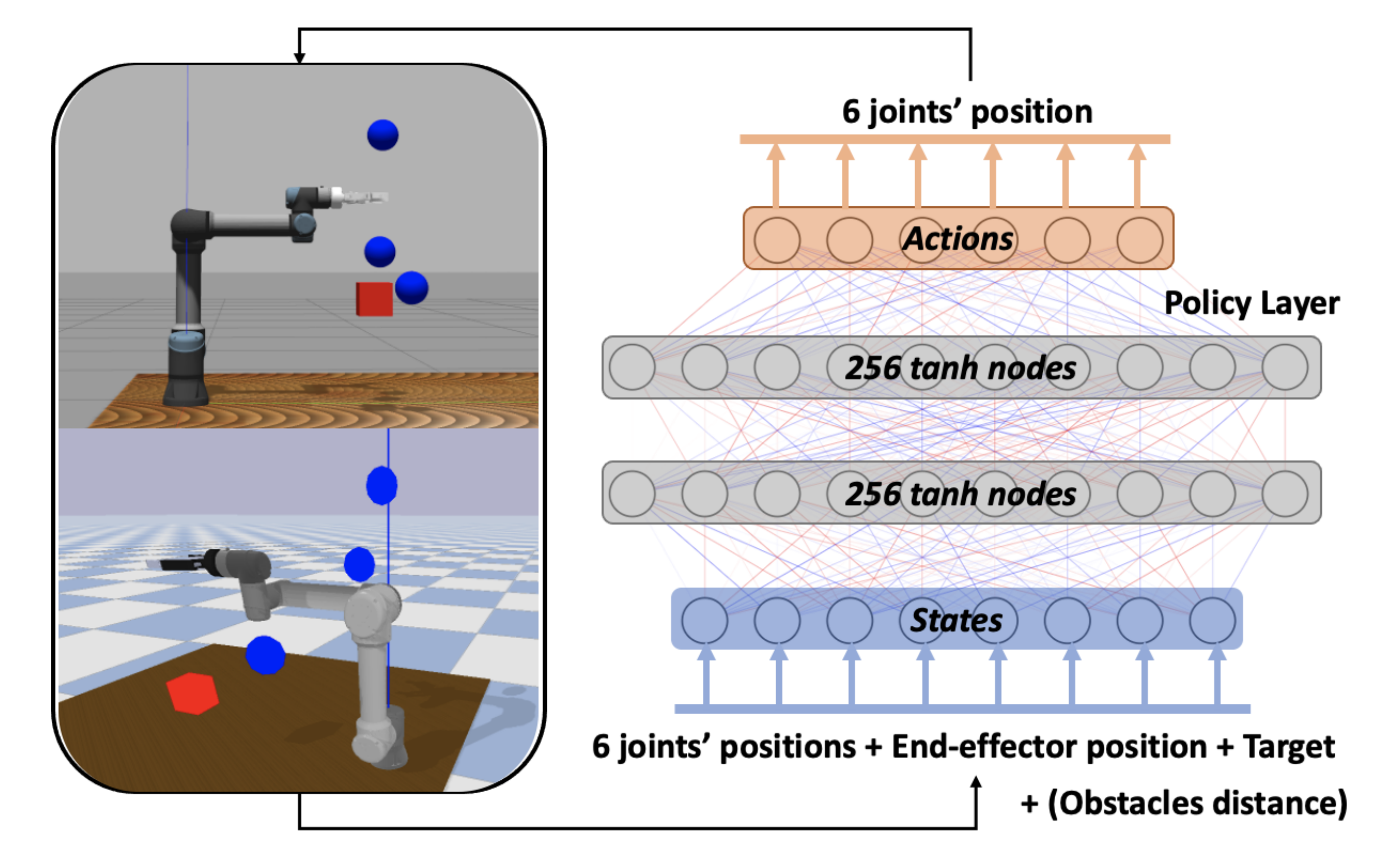}
      \vspace{-4mm}
      \caption{The framework of the strategy for learning: The state of the robot consists of the current joint angles $q_t$, the position of end-effector $p_e$, and the target position $p_t$, which extends with the distance of three obstacles when the task is with obstacles. The grey layers are the network structure. The action layer produces the desired joint angles. }
      \label{figure4}
  \end{figure}

In our system, both the Actor and Critic (AC) neural network consist of three layers, where each layer consists of $256$ neurons. The first two layers use the $tanh$ activation function. The only distinction between actor and critic networks is that the critic network generates only a single scalar value, while the actor produces a vector of six values, representing the robot joints' position. For both networks, we consider the Adam optimizer. The overall framework of our approach is depicted in Fig.~\ref{figure4}. 

\section{IMPROVED PROXIMAL POLICY OPTIMIZATION}
Original PPO does not perform well in solving complex robotic problems (i.e., the discussed reaching task while there are obstacles in the environment). In particular, we observed that PPO took too long time to be trained in the entire workspace, and the accuracy was poor for the reaching task, even without collision avoidance. To overcome these limitations, we propose two improvements for the PPO to achieve better results, which are discussed in detail in the following subsections.

\subsection{Action ensembles Based on Poisson Distribution} 

For most tasks that applied learning methods in robotics, Gaussian distribution is utilized to describe the optimal policy distribution, which is on the grounds that Gaussian distribution is more realistic. In the reaching task, the distribution of choosing action can be considered as $\pi_\theta(a_t|s_t)\sim N(\mu_\theta(s_t),\delta_\theta)$, where $\delta$ represents the uncertainty and unstable of distribution to output optimal action. To some extent, averaging the multiple outputs can solve this problem but it will limit the exploration ability at the initial steps and make the policy easily prone to local optima. 

To make the policy robust, balance the exploration and avoid inclining to optima, we select the number of samples through Poisson distribution~\cite{9636681}. In particular, the specific calculation is defined as:
\begin{equation}
    i\sim clip(Poisson(\beta), 1, \beta),\quad \beta = 1+ \alpha \dfrac{e_n}{e_a}
\end{equation}
\begin{equation}
    \bm a_{t,j}\sim N(\mu_\theta(s_t), \delta_\theta),\quad \bm a_t = \underset{j}{mean}(\bm a_{t,j})
\end{equation}
where $\beta$ indicates the Poisson distribution mean, $\alpha=12$, $j\in[1,j)$, $e_n$ and $e_a$ represent the number of episodes at the current episode and the final episode, respectively.

\subsection{AC Architecture with Policy Feedback}

PPO uses the standard AC architecture, which means that the critic network estimates the value function that the actor-network uses to improve policy performance. However, the policy does not participate in the update of the value function directly, which increases the instability of DRL algorithms. Thus, inspired by \cite{gu2021proximal}, we include the policy in the value function update. Using this strategy, the critic network can recognize policy differences rapidly. To put the strategy into action, we utilize an adaptive clipped discount factor:
\begin{equation}
    \gamma(s,a;\eta) = clip(\pi(s,a), \eta, 1) \quad \eta\in(0.6, 0.99)
\end{equation}
in which $\pi(s,a)$ represents the policy and the adaptive $\gamma$ can join in the update of critic network.

\subsection{Improved PPO}

\begin{figure}[!t]
  \centering
  \includegraphics[width=1.0\linewidth]{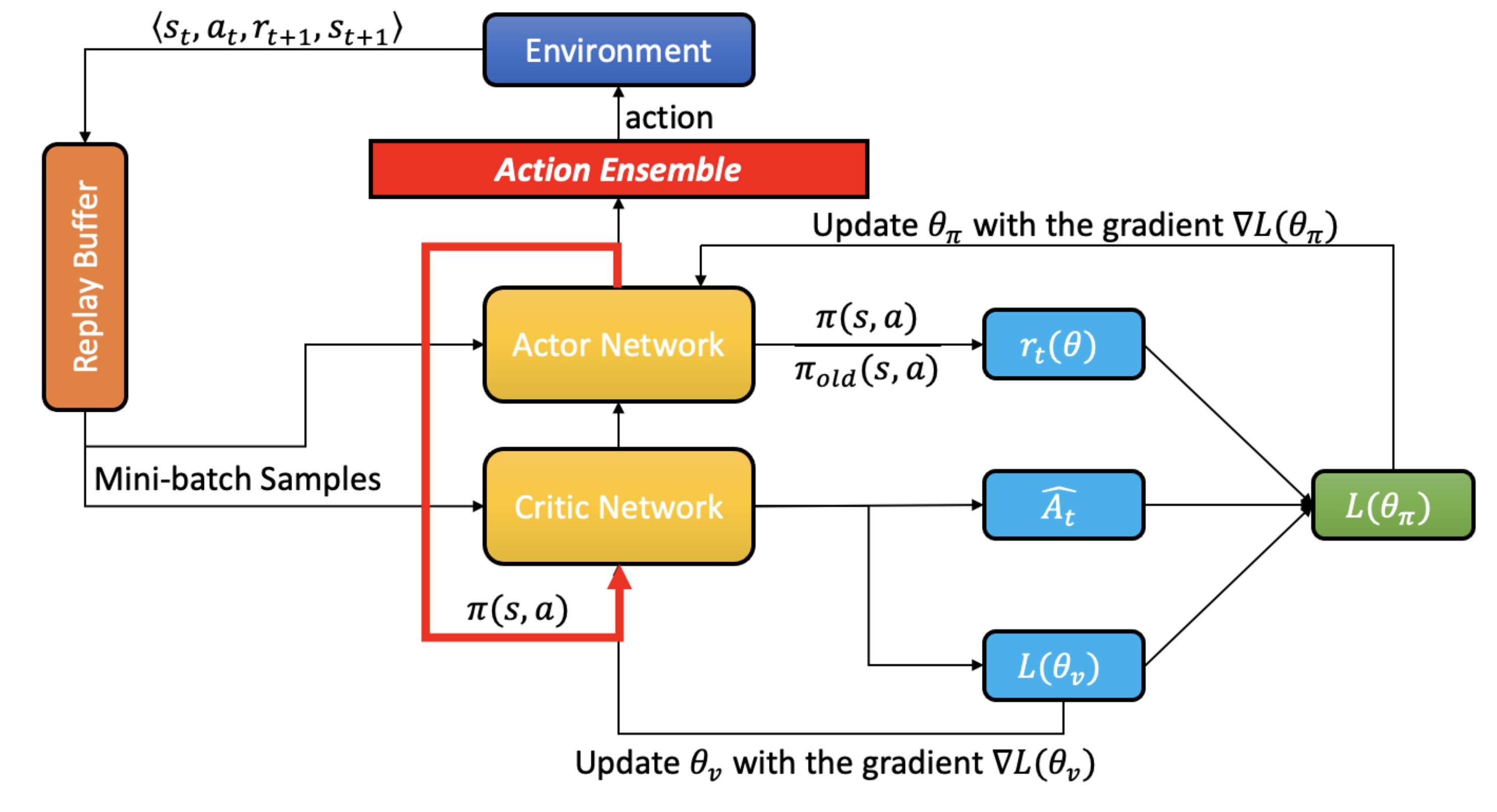}
  \vspace{-5mm}
  \caption{The framework of the improved PPO: The action ensemble (\textit{shown in red block}) and the policy feedback (\textit{red line}) are the improvements that we proposed.}
  \label{figure5}
\end{figure}

According to previous advancements, the loss functions of actor-critic networks can be represented as follows:
\begin{equation}
   \begin{aligned}
    \bm{L}(\theta^{\pi})&=\mathbb{\dot{E}}_t[min(r_t(\theta^{\pi}){A_t}, \\
   &\quad clip(r_t(\theta^{\pi}),1-\epsilon,1+\epsilon){A_t})]
   \end{aligned}
\end{equation}
\begin{equation}
    \bm{L}(\theta^{V})=\mathbb{\dot{E}}_t[(R_t^\pi-V_t)^2]
\end{equation}
Fig.~\ref{figure5} shows the overall architecture of the improved PPO, and Algorithm ~\ref{alg1} represents the pseudocode of the improved PPO.
By combining all the proposed strategies together, the entire algorithm for the robot to avoid obstacles while reaching the goal is summarized in Algorithm ~\ref{alg2}.

\begin{algorithm}[!t]
    \caption{Improved PPO Pseudocode}
    \label{alg1}
      Orthogonal initialize the actor and critic networks\\
      Initialize optimizer as Adam with learning rates\\
      Set $\lambda$, $a_{max} = 3.14$, clip parameter: $\epsilon$\\
      Set AEP parameter: $\alpha=12$\\
      Set Policy Feedback parameter: $\eta\in[0.6,0.99]$\\
      \While{True}{
         Get $s_t$ from the environment\\
         Sample $a_t$ from the baseline policy $\pi_{b\theta}(a_t|s_t)$\\ 
         Using $\bm a_t\sim AEP(\mu_\theta(s_t),\delta_\theta)$ to publish $\bm a_t$ \\
         Execute $a_t$, get $<s_t, a_t, \pi_\theta, r_t, s_{t+1}>$ and store them into buffer $D$\\
         Calculate clipped discount factor $\gamma(s,a;\eta)$ by Eq.(15)\\
         Compute reward: $R_t^\pi = \sum \limits_{i=t}^{T}r_i\prod \limits_{j=t}^n \gamma(s,a;\eta)$\\
         Compute advantage function: $A_t^\pi=r_t+\gamma V_{t+1}^\pi-V_t^\pi$\\
         Use Eq.(16) to accumulate gradients with respect to $\theta_\pi$\\
         Use Eq.(17) to accumulate gradients with respect to $\theta_V$
      } 
\end{algorithm}

\begin{algorithm}[!t]
    \caption{The Manipulator Control via Improved PPO}
    \label{alg2}
    \For{number of epochs}{
        Set i\_epoch as random seed\\
        Orthogonal initialize the actor and critic networks\\
        Set $\lambda$, maximum action, learning rates, clip parameter $\epsilon$ \\
        \For{number of episodes}{
        Set a target position $p_t$ randomly\\
        \For{numbers of maximum time steps}{
        Get $s_t$ from the environment\\
        Chose the action using $\bm a_t\sim AEP(\mu_\theta(s_t),\delta_\theta)$\\
        Clip $\bm a_t$ to ensure safety for the environment\\
        Execute $a_t$, get $<s_t, a_t, \pi_\theta, r_t, s_{t+1}>$ and store them into buffer $D$
        } 
    }
        \For{every time step t}{
        \For{$k$ epochs}{
        Select a minibatch $b_k$ in $D$\\
        Utilize Algorithm 1 to Update networks
    }
    }
    }
\end{algorithm}

\section{EXPERIMENTS}

To assess the performance of the proposed approach, we conducted a set of experiments in simulation and real-robot environments. The experiments were divided into four rounds. In the first round, we compared the proposed approach to six baseline methods and evaluated their performance based on the accumulated reward. In the second round, we evaluated the success rate of the goal-reaching task under both obstacle-free and obstacle-present conditions. These rounds of experiments were carried out in Pybullet simulation. In the next round of experiments, the trained model was transferred from Pybullet to Gazebo. In the last round of evaluation, we employed the learned policy in real-robot without any fine-tuning. We used the same network and code in both simulated and real experiments. All experiments were run on a PC with Ubuntu $20.04$, featuring a $3.20$ GHz Intel Xeon(R) i$7$ processor and an RTX $2080$ Ti NVIDIA graphics card.

\begin{figure*}[!t]
        \subfigure[]{
            \includegraphics[width=0.33\linewidth]{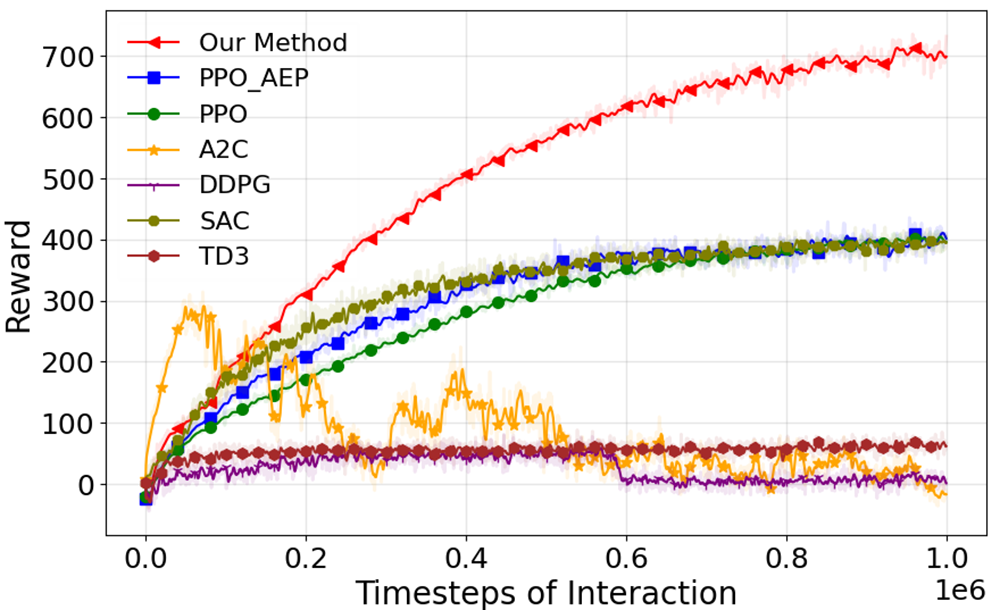}
        }
        \subfigure[]{
            \includegraphics[width=0.33\linewidth]{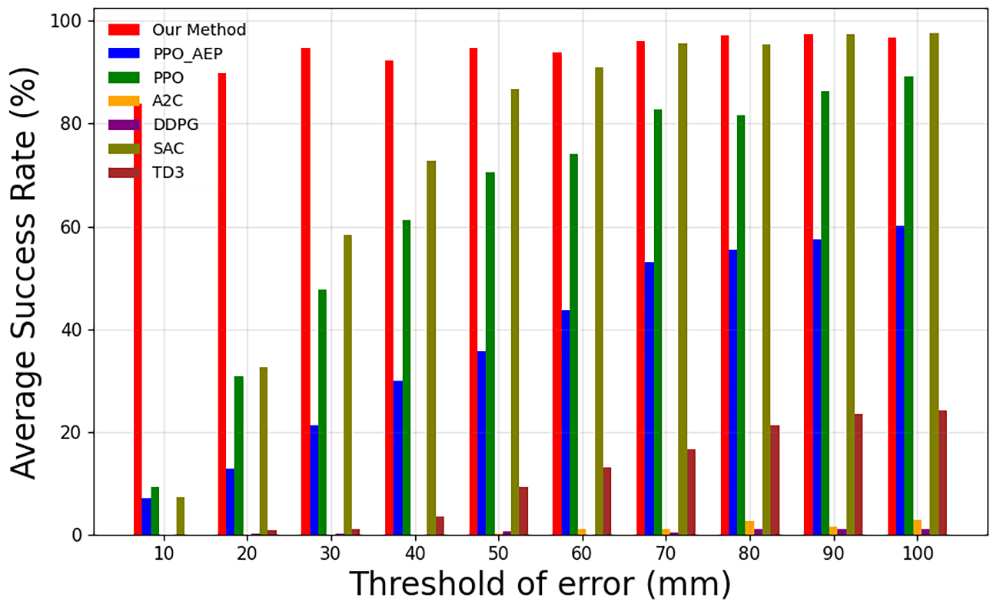}
        }
        \subfigure[]{
            \raisebox{0.07\height}{\includegraphics[width=0.30\linewidth]{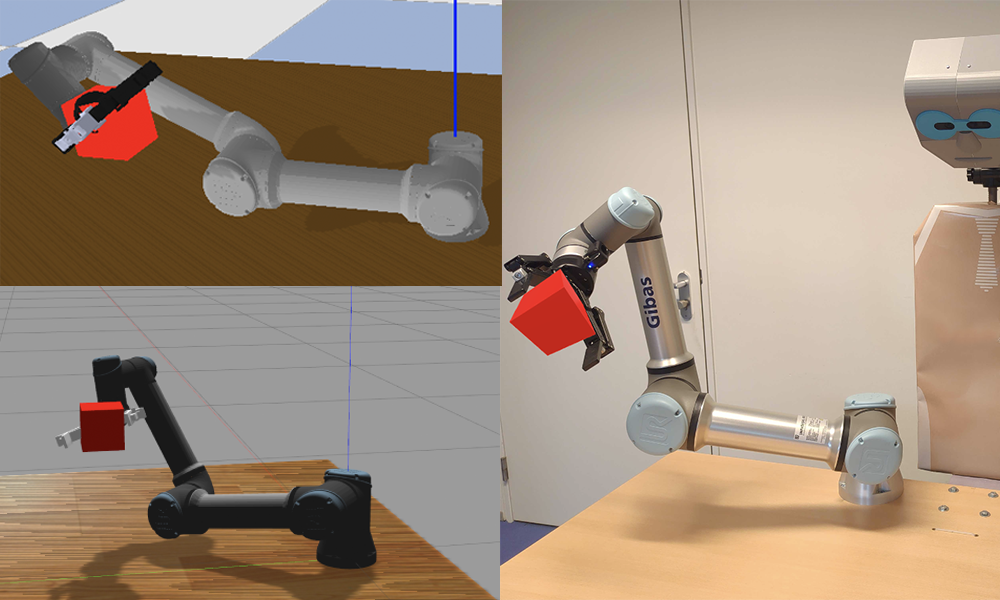}}
        }
        \subfigure[]{
            \includegraphics[width=0.33\linewidth]{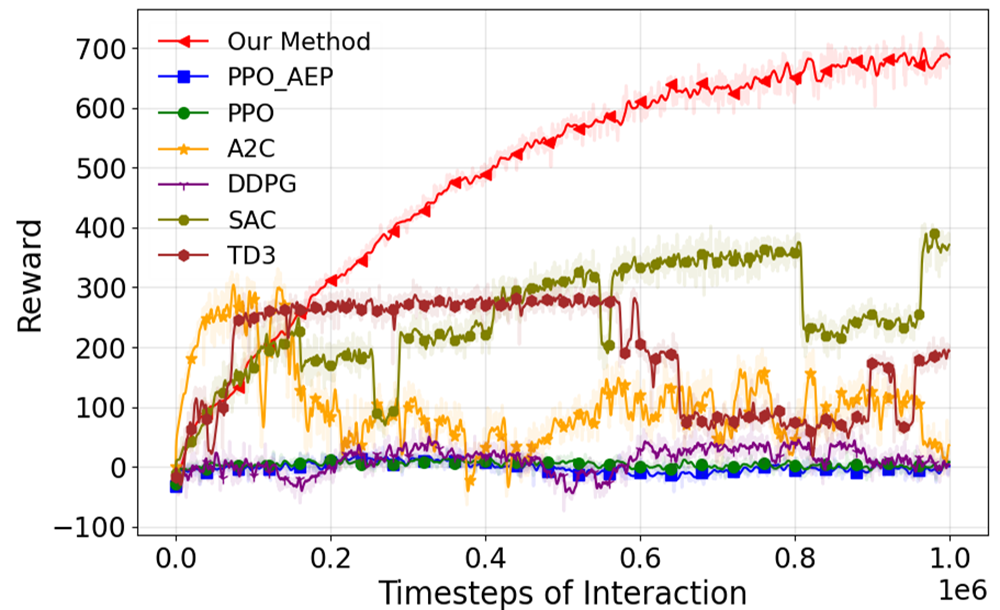}
        }
        \subfigure[]{
            \includegraphics[width=0.33\linewidth]{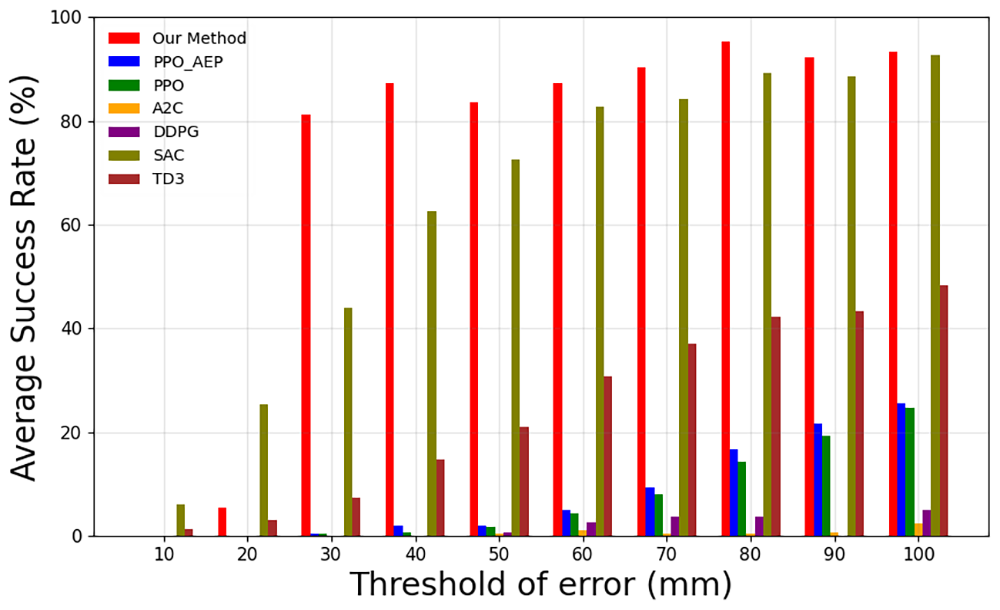}
        }
        \subfigure[]{
            \raisebox{0.07\height}{\includegraphics[width=0.30\linewidth]{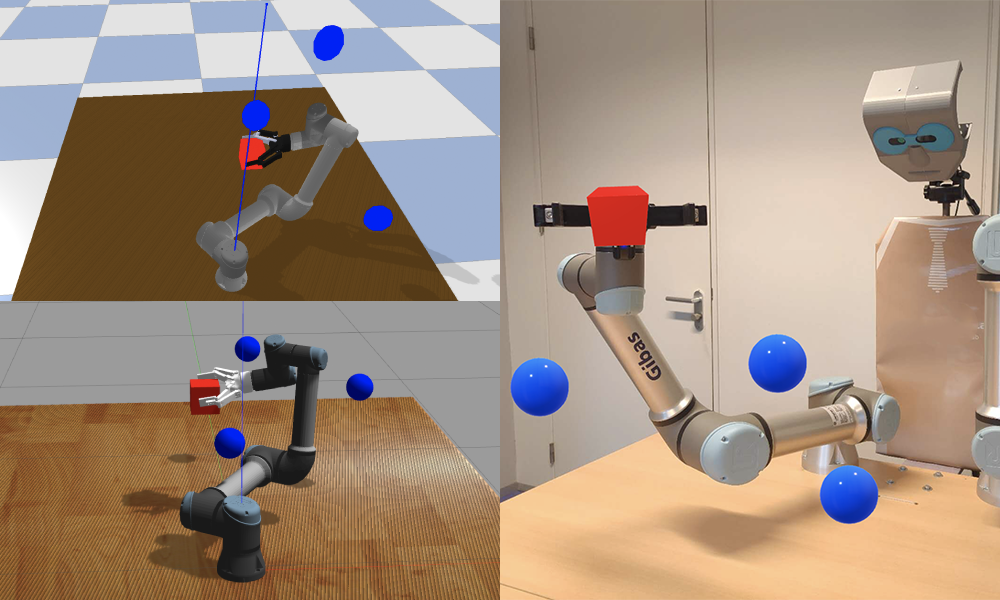}}
        }
    \caption{Summary of experiments: (a) Comparison result for reaching task without obstacles utilizing 5 random seeds; (b) Comparison of success rate on reaching task without obstacles; (c) The train and test simulation experiments for reaching task without obstacles in Pybullet, Gazebo and real robot experiments; (d) Comparison result for reaching task with obstacles utilizing 3 random seeds; (e) Comparison of success rate on reaching task with obstacles;  (f) The train and test simulation experiments for reaching task with obstacles in Pybullet, Gazebo and real robot experiments.}   
	\label{figure6}  
\end{figure*}

\subsection{Experimental setup}

\begin{figure}[!t]
  \centering  \includegraphics[width=1.0\linewidth]{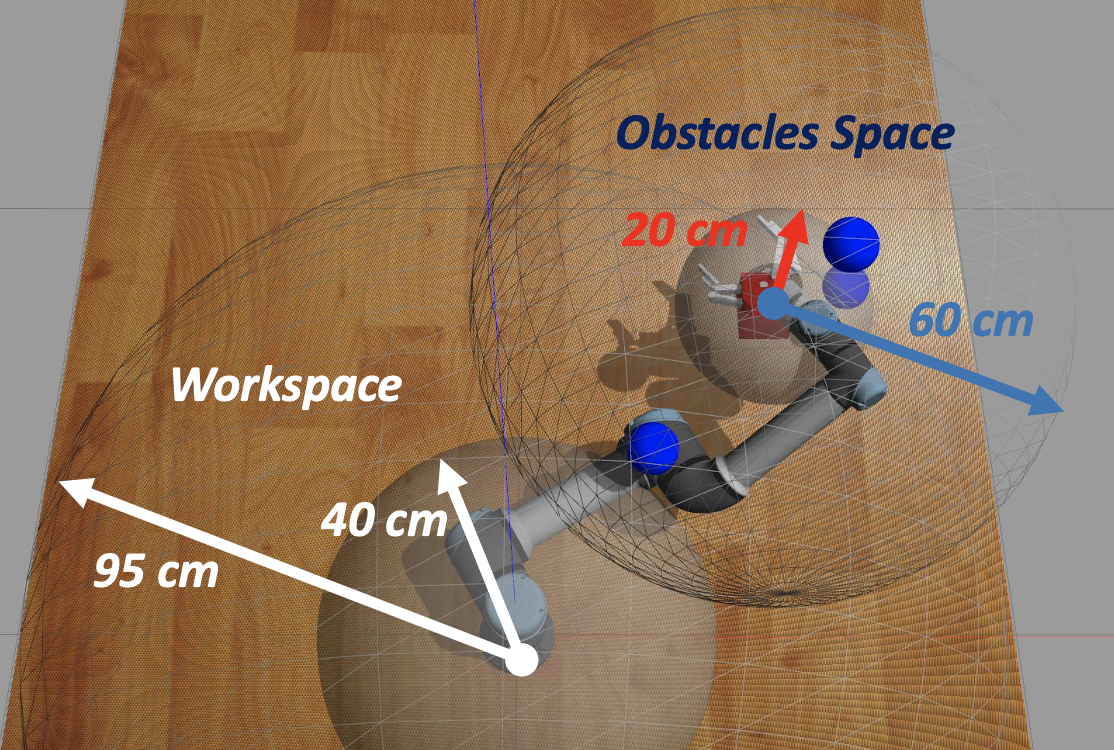}
  \vspace{-5mm}
  \caption{The workspace of the robot for reaching tasks with/without obstacles in Gazebo: the goal pose is shown by a red cube and the obstacle is shown by a blue sphere.}
  \label{figure8}
\end{figure}

As depicted in Fig.~\ref{figure8}, the workspace for the experiment is defined as a quarter spherical annulus with a major radius of $95$cm and a minor radius of $40$cm, centered at the base of the robot. The target position is randomly placed within the defined workspace during training. To increase the challenge of the reaching task, three spherical obstacles with a diameter of $0.05$m were placed around the target or near the manipulator link. The positions of the obstacles are randomly generated in a quarter spherical annulus with a major radius of $60$cm and a minor radius of $20$cm, centered at the target position. Obstacle positions that fall under the table are regenerated. The total number of episodes is $10^4$ and the maximum number of time steps for each episode is $100$, resulting in a total of $1$ million time steps. To ensure the validity of the results, all methods were trained five times with five different random seeds for the reaching task with and without obstacles. The termination condition of each episode during training is the distance between the target and the end-effector being less than $0.1$cm. During testing, $20$ different termination conditions ranging from $0.5$cm to $10$cm are set as a threshold to assess the success rate of reaching with and without obstacles, and $100$ targets are assigned at random for each threshold. To prevent collisions between the manipulator and the table, as well as self-collisions among all joints during the test, constraints are established for six joints.

\subsection{Sim-to-Sim and Sim-to-Real transfer}


The performance of the improved PPO was evaluated through two sets of experiments, one in the Gazebo environment and the other in the Pybullet environment, both lasting for $10^6$ time steps. The time spent on training is summarized in Table~\ref{table:time}, and it is evident that training in Pybullet was significantly faster compared to training in Gazebo. The accuracy of the reaching task was assessed by conducting $3000$ obstacle-free trials with the same model. Results, shown in Fig.~\ref{figure9}, indicate that the average error was $6 \pm 2$ mm. The trained model in Pybullet was also transferred to Gazebo to examine the feasibility of Sim-to-Sim transfer. Further experiments were carried out with $3000$ obstacle-free trials and the results were compared to those obtained from the model trained in Pybullet. The results showed that the agent performed equally well in both environments with a distance error between the target and end-effector of $\pm 0.5$ mm. This experiment demonstrated the benefits of Sim-to-Sim transfer in terms of reducing training time while preserving accuracy in the reaching task.

\begin{figure*}[!t]
\includegraphics[width=0.24\linewidth]{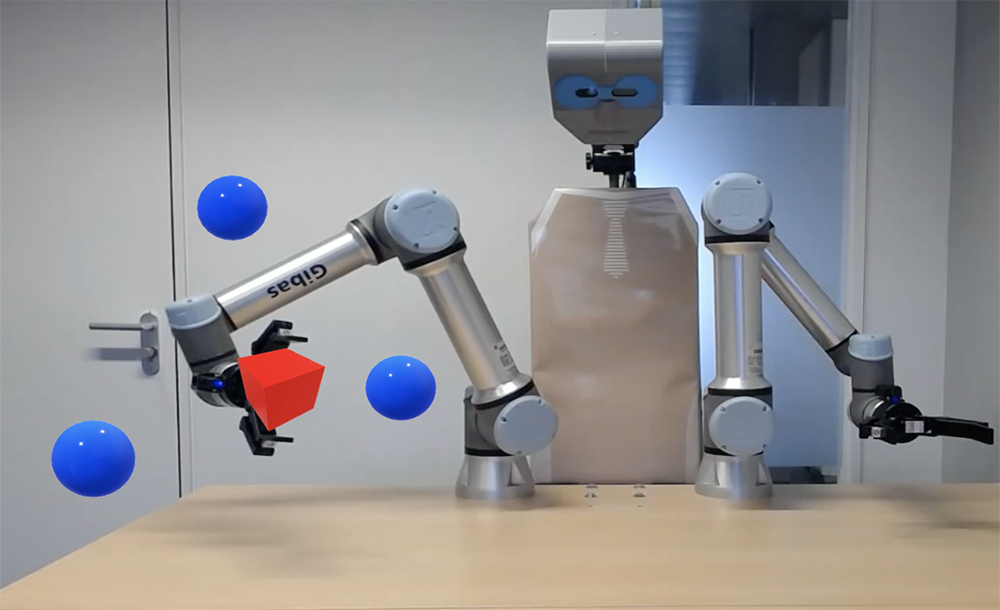}
\includegraphics[width=0.24\linewidth]{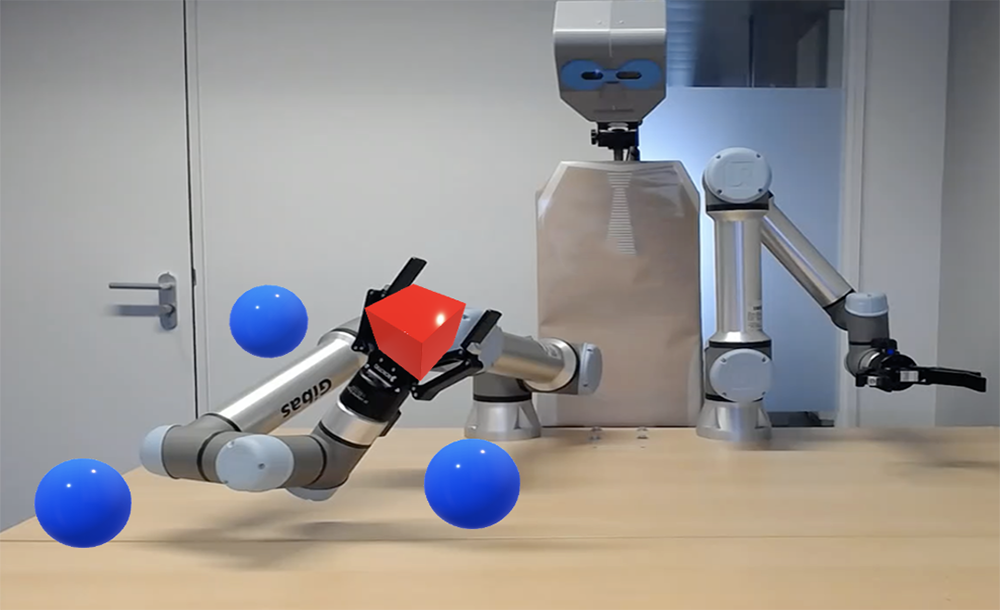}
\includegraphics[width=0.24\linewidth]{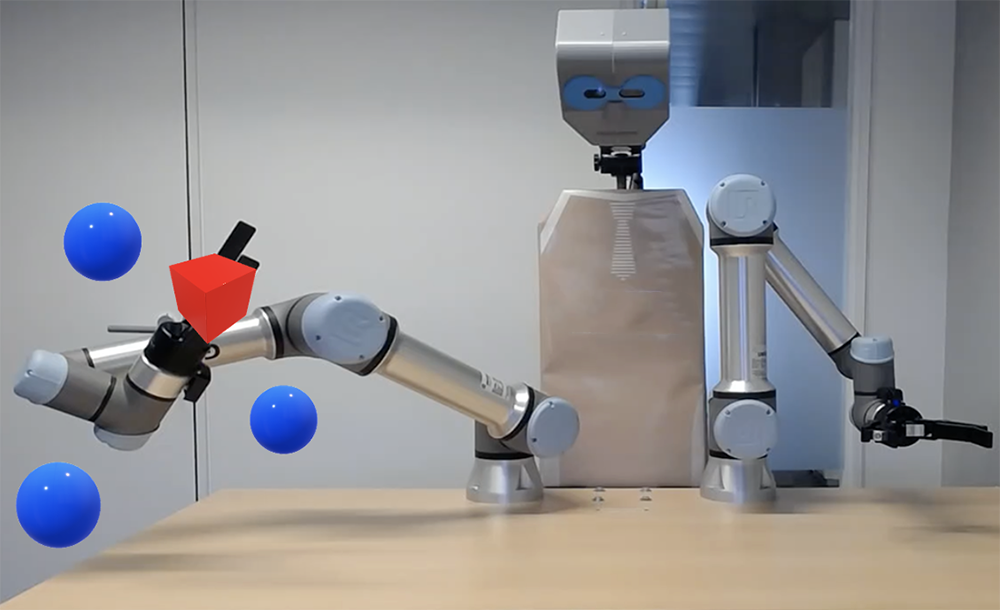}
\includegraphics[width=0.24\linewidth]{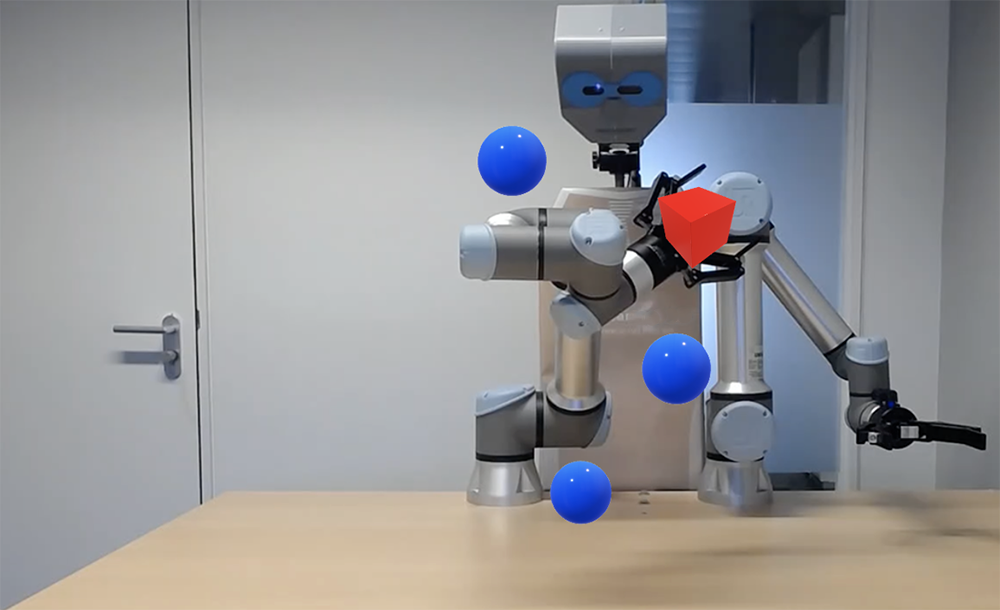}
    \caption{A series of four snapshots demonstrating the robot could successfully execute the reaching task while avoiding three obstacles in a real-world scenario. These figures showcase the effective application of the proposed method in navigating through cluttered environments. }   
	\label{figure12}  
\end{figure*}


To validate the performance of the proposed method for sim-to-real transfer, 50 trials for goal-reaching tasks in both the presence and absence of obstacles were conducted using a physical robot. The results are depicted in Fig.\ref{figure6} (\textit{c} and \textit{f}) and Fig.\ref{figure12}. The results demonstrate that the algorithm was effective in real-world scenarios to the same extent as it was in the simulation. A video summarizing these experiments is provided as supplementary material to the paper.

\begin{table}[!t]
\caption{\textbf{Training time comparison}}
\centering
\begin{tabular}{cccc}
\toprule
Task & Time steps & Gazebo & Pybullet \\
\midrule
Reaching without obstacles & $10^6$ & 50 h 06 min & 2 h 40 min \\
Reaching with obstacles & $10^6$ & $\ast \ast$ & 11 h 28 min \\
\bottomrule
\label{table:time}
\vspace{-4mm}
\end{tabular}
\end{table}

\begin{figure}[!b]
  \centering
  \includegraphics[width=0.9\linewidth]{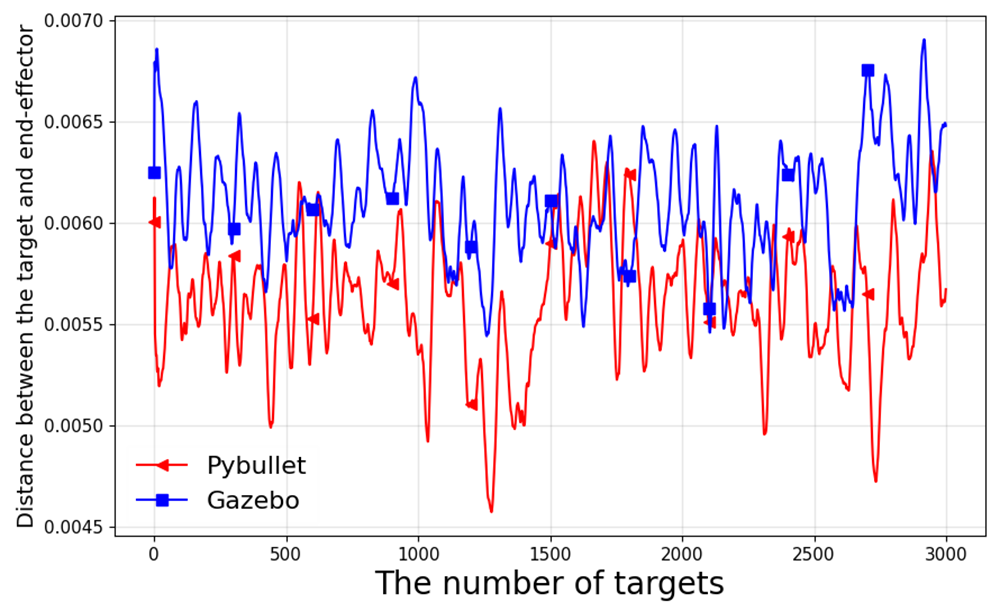}
  \caption{The distance between the target and end-effector for reaching task without obstacles in Gazebo and Pybullet using the same trained model obtained from Pybullet.}
  \label{figure9}
\end{figure}


\subsection{Comparison with baseline methods}
To demonstrate the superiority of our approach, we compared it with six state-of-the-art algorithms. We chose the following methods as the baselines: PPO-AEP \cite{9636681}, PPO \cite{schulman2017proximal}, Advantage Actor-Critic (A2C)\cite{kwon2017a2c}, Deep Deterministic Policy Gradient (DDPG) \cite{lillicrap2015continuous}, Soft Actor-Critic (SAC) \cite{haarnoja2018soft} and Twin Delayed DDPG (TD3) \cite{dankwa2019twin}. 
Fig.~\ref{figure6} showed the comparison between the proposed method with the baseline methods. It should be noted that the baselines used the same neural network, parameters, state-action representation, and reward function. The learning curves, as shown in Fig.~\ref{figure6} (a) and (d), indicated that our method could get the highest reward in the same condition. Fig.~\ref{figure6} (b) and (e) also illustrated that the success rate of our method reached the highest value in the reaching task with and without obstacles especially when the threshold is set below $50$ mm. 

\subsubsection{Reaching Task without obstacle}

For observing the performance of the reaching task without obstacle avoidance, the success rate was produced by recording the number of times the robot could reach the target pose within a predefined threshold, ranging from $10$mm to $100$mm with the interval of $10$mm. For each of the threshold values, we performed 100 experiments by randomly setting the target point in each experiment. We repeated such experiments with five random seeds and reported the average success rate for each threshold value. The obtained results were summarized in Fig.~\ref{figure6} (b). By comparing the results, it was visible that our approach outperformed the other baselines. Meanwhile, by setting the threshold to $50$mm, our method achieved a 100 percent success rate. In particular, even if the threshold was below $10$mm, our method still could achieve over 65 percent success rate, which was useful for precise control and path planning in the future.  

\begin{figure}[!t]
  \centering
  \includegraphics[width=1.0\linewidth]{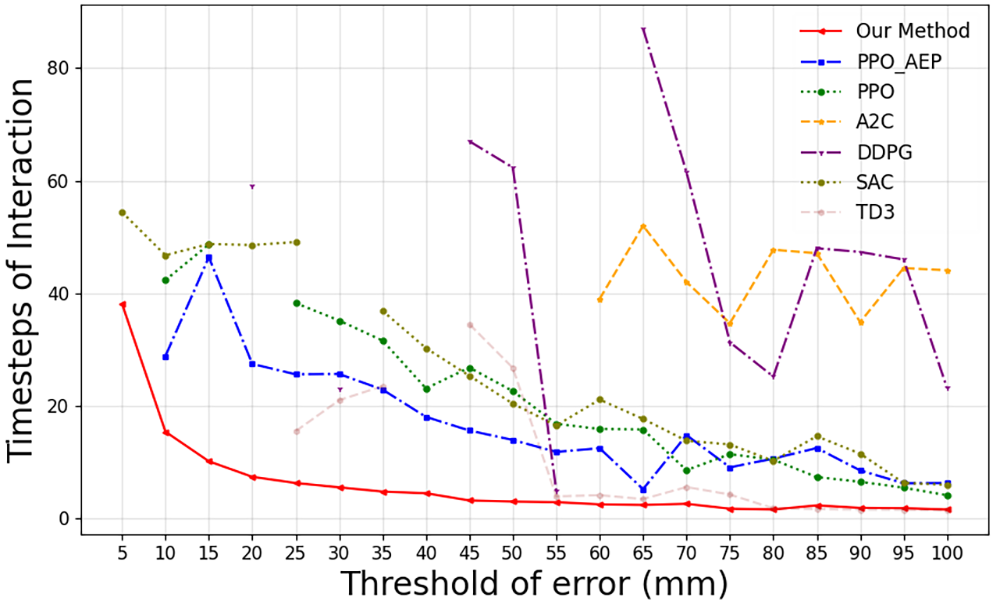}
  \vspace{-4mm}
  \caption{The timesteps for reaching task without obstacles in test experiments obtained from Pybullet: by comparing all the approaches, it is visible that our approach reaches the target goal faster than the other approaches.}
   \label{figure11}
\end{figure}

We also compared all the approaches in terms of how fast they can reach the goal as a function of the timesteps versus the threshold of the error. 
Results are summarized in Fig.~\ref{figure11}.  It can be observed that the proposed method achieved the goal in the minimum number of timesteps, whereas DDPG and A2C required a relatively larger number of timesteps to reach the target. These results indicate that DDPG and A2C are not appropriate for complex high-dimensional learning tasks.

\subsubsection{Reaching task while avoiding obstacle }
The learning curve of all methods in reaching tasks with obstacles is shown in Fig.~\ref{figure6} (d). By comparing the results, it is visible that our approach outperformed the other baselines by a large margin. Fig.~\ref{figure6} (e) summarized the average success rate obtained by all methods for this round of experiments. By comparing the results, it is visible that the agent with the proposed IPPO and SAC policies demonstrated superior performance in comparison to the other methods. Specifically, when the error threshold was set above $20$mm, the IPPO policy exhibited the highest success rate, whereas, for error thresholds less than $20$mm, the SAC policy demonstrated a better success rate than the other methods. These results emphasize the effectiveness of the proposed IPPO and SAC policies in complex high-dimensional learning tasks.
We hypothesize that the underlying reason for the superior performance of IPPO and SAC compared to DDPG, A2C, and TD3 is likely due to the use of  actor-critic methods and incorporation of entropy into the optimization process, which would lead to better exploration-exploitation trade-offs and stability during training. Additionally, the proposed IPPO uses an action ensemble, which allows for faster convergence and better performance compared to other algorithms. On the other hand, DDPG, A2C, and TD3 are based on traditional reinforcement learning algorithms that may not be able to handle high-dimensional and complex environments.




\section{Conclusion} 
\label{sec:conclusion}

In this paper, we proposed an effective deep reinforcement learning method, called IPPO, for joint-level obstacle avoidance in manipulators. The method was implemented as a multi-layer neural network optimized by an improved version of the Proximal Policy Optimization algorithm. In particular, we designed a special calculation for the shortest distance between obstacles and links of the manipulator and also incorporated two improvements to the original PPO. Both simulation and real-robot experiments were conducted to validate the performance of the proposed method through Sim-to-Sim and Sim-to-Real transfer techniques. We compared the performance of the proposed approach with six RL baselines. Experimental results showed that the proposed method was efficient for reaching tasks with/without obstacles and could outperform the selected baselines by a large margin in terms of accumulated rewards, success rate, and the number of timesteps to reach to the target goal. In the continuation of this work, we would like to look into the feasibility of using the proposed approach in dynamic scenarios where the pose of the target and the obstacles changes over time.


\bibliographystyle{ieeetr}
\bibliography{references}

\end{document}